%% file: Analyzing_Hypothesis-Only_Biases_Removal_in_NLI_Models_arxiv.tex
\documentclass[11pt,a4paper]{article}
\usepackage[hyperref]{naaclhlt2019}
\usepackage{times}
\usepackage{latexsym}

\input{math_commands.tex}

\usepackage{soul,color}
\usepackage{hyperref}
\usepackage{url}

\usepackage{subcaption}
\usepackage{amsmath}
\usepackage{amssymb}
\usepackage{comment}
\usepackage{booktabs}
\usepackage{tabularx}
\usepackage{dcolumn}
\newcolumntype{d}[1]{D{.}{\cdot}{#1} }
\usepackage[normalem]{ulem}

\usepackage{adjustbox}

\usepackage{dcolumn}
\newcolumntype{H}{>{\setbox0=\hbox\bgroup}c<{\egroup}@{}}

\usepackage[]{todonotes}

\usepackage{xspace}
\newcommand{\hypoth}{H\xspace}
\newcommand{\premise}{P\xspace}
\newcommand{\enc}{g}

\newcommand{\vV}{\vv}
\newcommand{\vP}{\vp}
\newcommand{\vH}{\vh}
\newcommand{\classifier}{c}
\newcommand{\lY}{y}

\newcommand{\HOC}{AdvCls\xspace}
\newcommand{\NS}{AdvDat\xspace}

\newcommand{\newsecref}[1]{\S\ref{#1}}
\newcommand{\newssecref}[1]{Subsection~\ref{#1}}
\newcommand{\newfigref}[1]{Figure~\ref{#1}}
\newcommand{\tabref}[1]{Table~\ref{#1}}
\newcommand{\appref}[1]{Appendix~\ref{#1}}

\usepackage{caption}

\usepackage{paralist}
\usepackage{enumitem}

\title{
On Adversarial Removal of Hypothesis-only Bias in \\Natural Language Inference
}

\author{Yonatan Belinkov$^{13}$\thanks{$^*$ Equal contribution} \hspace{1em} Adam Poliak$^{2*}$ \\
\textbf{\hspace{1em} Stuart M. Shieber$^1$ \hspace{1em} Benjamin Van Durme$^2$ \hspace{1em} Alexander M. Rush$^1$}\\
$^1$Harvard University \hspace{0.9em} 
$^2$Johns Hopkins University \hspace{0.9em} 
$^3$Massachusetts Institute of Technology \\
\texttt{\{belinkov,shieber,srush\}@seas.harvard.edu} \\ 
\texttt{\{azpoliak,vandurme\}@cs.jhu.edu}
}

\newcommand{\cameraready}[1]{\xspace}
\renewcommand{\cameraready}[1]{\xspace}
\newcommand{\adversarialDescription}[1]{}
\renewcommand{\adversarialDescription}[1]{}
\newcommand{\jointDescription}[1]{}
\renewcommand{\jointDescription}[1]{}

\aclfinalcopy 
\def\aclpaperid{34} 
\begin{document}

\maketitle

\begin{abstract}
Popular Natural Language Inference
(NLI) datasets have been shown to
be tainted by hypothesis-only biases.
Adversarial learning 
may help
models 
ignore sensitive
biases and spurious correlations in data. 
We evaluate
whether adversarial learning
can be used in NLI to encourage
models to learn
representations free of hypothesis-only biases.
Our 
analyses indicate that the
representations learned
via adversarial learning may be
less biased,
with only small drops in NLI accuracy.
\end{abstract}

\section{Introduction}
Popular datasets for Natural Language Inference (NLI) 
- the task of determining
whether one sentence (premise) likely entails another (hypothesis) - 
contain hypothesis-only biases that allow models to perform the task surprisingly well by
only considering hypotheses while ignoring the corresponding premises.
For 
instance, such a method
correctly predicted the examples in \tabref{tab:hyp-only-contradictions} as contradictions.
As datasets may always contain biases,
it is important to 
analyze whether, and to what extent, models are immune to or rely on known biases.
Furthermore, it is important to build models
that can overcome these biases.

Recent work in NLP aims to build more robust systems using adversarial methods~\cite[][\textit{i.a.}]{alzantot2018generating,N18-1111,belinkov2018synthetic}.
In particular, \citet{elazar2018adversarial} attempted to use adversarial training to remove demographic attributes from text data, with limited success.
Inspired by this line of work,
we use adversarial learning
to add small components to an existing and popular NLI system that has been
used to learn general sentence representations~\cite{D17-1070}.
The adversarial techniques include (1) using an external adversarial classifier conditioned on  hypotheses alone, and (2) creating noisy, perturbed training examples.
In our analyses we ask
whether hidden, hypothesis-only biases are no longer present in the resulting sentence representations
after adversarial learning.
The goal is to build models with less bias, ideally while limiting the inevitable degradation in task performance.
Our results suggest that progress on this goal
may depend on which adversarial learning techniques are
used.

Although recent work has applied adversarial learning to NLI~\cite{minerviniconll18,kang-EtAl:2018:Long}, this is the first work to our knowledge
that explicitly studies NLI models designed to ignore hypothesis-only biases.

\begin{table}[t!]
\centering
\begin{tabular}{l}
\toprule
A dog runs through the woods near a cottage \\
$\blacktriangleright$  The dog is \textit{sleeping} on the ground   \\ \midrule
A person writing something on a newspaper \\
$\blacktriangleright$ A person is \textit{driving} a fire truck \\ \midrule
A man is doing tricks on a skateboard \\
$\blacktriangleright$ \textit{Nobody} is doing tricks \\
\bottomrule
\end{tabular}
\caption{Examples from SNLI's development set that \citet{poliak-EtAl:2018:S18-2}'s hypothesis-only model correctly predicted as contradictions. The first line in each section is a premise and 
lines with $\blacktriangleright$ are corresponding hypotheses. The italicized words are correlated with the ``contradiction'' label
in SNLI}
\label{tab:hyp-only-contradictions}
\end{table}

\section{Methods}
\label{sec:methods}

We consider two types of adversarial methods. 
In the first method,
we incorporate
an external 
classifier to force
the hypothesis-encoder to ignore hypothesis-only biases.
In the second method, we randomly swap premises in the training set to create noisy examples.

\subsection{
General NLI Model}
Let $(\premise, \hypoth)$ denote a premise-hypothesis pair, 
$\enc$
denote an encoder that maps a sentence $S$ to a vector representation $\vv$,
and $\classifier$
a classifier that maps 
$\vV$ to an output label $\lY$.
A general NLI framework
contains the following components:

\begin{itemize}
\vspace{-5pt}
\item A \textbf{premise encoder} $\enc_P$ that maps the premise $\premise$ to a vector representation $\vP$.
\item A \textbf{hypothesis encoder} $\enc_H$ that maps the hypothesis $\hypoth$ to a vector representation $\vH$.
\item A \textbf{classifier} $\classifier_{\text{NLI}}$ that
combines and maps $\vP$ and $\vH$
to an
output $\lY$.
\end{itemize}
\vspace{-5pt}
In this model,
the premise and hypothesis are each encoded with separate encoders.
The NLI classifier is usually trained to minimize the 
objective:
\begin{equation}
L_\text{NLI} =
L(\classifier_\text{NLI}( [\enc_P(P) ; \enc_H(H)], y ))
\label{eq:normal-loss}
\end{equation}
where $L(\tilde{y}, y)$ is the cross-entropy loss.
If $\enc_P$
is not used, a model should
not
be able to successfully perform NLI. However, models without $\enc_P$ may achieve non-trivial results, indicating the existence of biases in 
hypotheses~\citep{gururangan-EtAl:2018:N18-2,poliak-EtAl:2018:S18-2,1804.08117}.

\subsection{
\HOC: Adversarial Classifier}

Our 
first approach, referred to as 
\HOC, follows 
the common adversarial training method~\citep{goodfellow2014explaining,ganin2015unsupervised,xie2017controllable,
zhang2018mitigating} by adding an additional adversarial classifier $\classifier_{\text{Hypoth}}$ to our model. $\classifier_{\text{Hypoth}}$ maps the hypothesis representation $\vH$ to an output
$\lY$.
In domain adversarial learning, the classifier is typically used to predict unwanted features, e.g., protected attributes like race, age, or gender~\cite{elazar2018adversarial}. Here,
we do not have explicit 
protected attributes but rather \emph{latent} 
hypothesis-only biases. Therefore,
we use $\classifier_\text{Hypoth}$ to predict the NLI label given only the hypothesis.  To successfully perform this prediction, $\classifier_\text{Hypoth}$  needs to exploit latent biases in 
$\vH$.

We modify the 
objective function (\ref{eq:normal-loss}) as
\begin{align*}
L =& L_\text{NLI} + \lambda_\text{Loss} L_\text{Adv} \\
L_\text{Adv} =&
L(\classifier_\text{Hypoth}(\lambda_\text{Enc}\text{GRL}_\lambda(\enc_H(H)), y))
\end{align*}
To control the interplay between $\classifier_\text{NLI}$ and $\classifier_\text{Hypoth}$ we set two hyper-parameters:
$\lambda_\text{Loss}$,
the importance of the adversarial loss function, and
$\lambda_\text{Enc}$,
a scaling factor that multiplies
the gradients after reversing them. This is implemented by the scaled gradient reversal layer, $\text{GRL}_\lambda$~\cite{ganin2015unsupervised}.
The goal here is modify the representation $\enc_H(H)$ so that it is maximally informative
for NLI while simultaneously minimizes the ability of $\classifier_\text{Hypoth}$ to accurately predict the NLI label.

\subsection{
\NS: Adversarial Training Data}

For our second approach, which we call \NS,
we use an unchanged general model, but train it with perturbed training data. 
For a fraction of example $(\premise,\hypoth)$ pairs in the training data, 
we replace $\premise$ 
with $\premise'$, a premise from another training example, chosen uniformly at random. For these instances, during back-propagation, we similarly reverse the gradient but only back-propagate through $\enc_{H}$. The adversarial loss function $L_\text{RandAdv}$
is defined as:
\begin{equation*}
L(\classifier_\text{NLI}( [\text{GRL}_0(\enc_P(P')) ; \lambda_\text{Enc}\text{GRL}_\lambda(\enc_H(H))], y ))
\end{equation*}
where
$\text{GRL}_0$ implements gradient blocking on
$\enc_P$ by using the identity function in the forward step and a zero gradient during the backward step. At the same time, $\text{GRL}_\lambda$ reverses the gradient going into $\enc_H$ 
and scales it by $\lambda_\text{Enc}$, as before.

We set a hyper-parameter $\lambda_\text{Rand} \in [0,1] $ that controls what fraction $\premise$'s are swapped at random. 
In turn, the final loss function combines the two 
losses 
based on $\lambda_\text{Rand}$ as
\begin{equation*}
L = (1-\lambda_\text{Rand}) L_\text{NLI} + \lambda_\text{Rand}L_\text{RandAdv}
\end{equation*}
In essence, this method penalizes the model for correctly predicting
$y$ in perturbed examples where the premise is uninformative.
This implicitly assumes that the label for ($\premise,\hypoth$)
should be different than the label for ($\premise',\hypoth$),
which in practice does not always hold true.\footnote{As pointed out by a reviewer,  a pair  labeled as neutral might end up remaining neutral after randomly sampling the premise, so adversarially training in this case might weaken the model. Instead, one could limit adversarial training to cases of entailment or contradiction.}

\section{Experiments
\& Results}

\paragraph{Experimental setup}
Out of
$10$ NLI datasets,
\citet{poliak-EtAl:2018:S18-2} found that the Stanford Natural Language Inference dataset~\cite[SNLI;][]{snli:emnlp2015} contained the most (or worst) hypothesis-only biases
---their hypothesis-only model outperformed the majority baseline by roughly $100\%$ (going from roughly $34\%$ to $69\%$). Because of
the large magnitude
of
these biases, confirmed by~\citet{1804.08117} and \citet{gururangan-EtAl:2018:N18-2},
we focus on SNLI.
We use the standard SNLI split and report validation and test results.
We also test 
on SNLI-hard, a subset of SNLI that~\citet{gururangan-EtAl:2018:N18-2} filtered such that it 
may not contain unwanted artifacts. 

We apply both adversarial techniques to
InferSent~\cite{D17-1070}, which serves as our general NLI architecture.\footnote{Code developed is available at \url{https://github.com/azpoliak/robust-nli}.}
Following the standard training details used in InferSent, 
we encode premises and hypotheses separately using bi-directional long short-term memory (BiLSTM) networks
~\cite{hochreiter1997long}.
Premises and hypotheses are initially mapped
(token-by-token) to Glove~\cite{pennington2014glove} representations. We use max-pooling over the BiLSTM states to extract premise and hypothesis representations and, following~\citet{mou-EtAl:2016:P16-2}, combine 
the representations by
concatenating their vectors, their difference, 
and their multiplication (element-wise).

We use the default training hyper-parameters in the released
InferSent codebase.\footnote{\url{https://github.com/facebookresearch/InferSent}}
These include setting the initial learning rate to $0.1$ and the decay rate to $0.99$,
using SGD optimization and
dividing the learning rate by $5$ at every epoch when the accuracy deceases on the validation set. 
The default settings also include stopping training
either when the learning rate drops below $10^{-5}$ or after $20$ epochs.
In both adversarial settings, the hyper-parameters are swept through $\{0.05, 0.1, 0.2, 0.4, 0.8, 1.0\}$.

\paragraph{Results}
~\tabref{tab:results-snli-basic} reports 
the results 
on SNLI, 
with the configurations that performed best on the validation set for each of the adversarial methods. 

\begin{table}[h]
\begin{center}
\begin{tabular}{l HrrrH}
\toprule
Model & Train & Val & Test & Hard & GLUE \\
\midrule
Baseline
& & 84.25 & 84.22 & 68.02 & \\
\midrule
\HOC & & 84.58 & 83.56 & 66.27 &  \\
\NS & & 78.45 & 78.30 & 55.60 & \\
\bottomrule
\end{tabular}
\end{center}
\vspace{-10pt}
\caption{Accuracies for the 
approaches.
Baseline refers to the unmodified, non-adversarial InferSent.}
\label{tab:results-snli-basic}
\end{table}

As expected, 
both training methods perform worse than our unmodified, non-adversarial InferSent baseline
on SNLI's test set,
since they remove biases that may be useful for performing this task. 
The difference for \HOC is minimal, and it even slightly outperforms InferSent on the validation set.  While \NS's results are noticeably lower than the 
non-adversarial InferSent,
the drops are still less than $6$\%
points.
\footnote{This drop may indicate that SNLI-hard may still have biases, but, as pointed out by a reviewer, an alternative explanation is a general loss of information in the encoded hypothesis. However,
\newssecref{sec:indicators} suggests that the loss of information is more focused on biases.}

\section{Analysis}

Our goal is to determine whether adversarial learning
can help 
build
NLI models without 
hypothesis-only biases. We first 
ask whether the models' learned sentence
representations can 
be used by a hypothesis-only classifier to 
perform well. We then explore the effects of increasing the adversarial strength, 
and end with a discussion of indicator words associated with hypothesis-only biases.

\begin{figure}[t!]
\centering
\begin{subfigure}[t]{\linewidth}
\centering
\includegraphics[width=\linewidth,trim={0 0 0 1.5cm},clip]{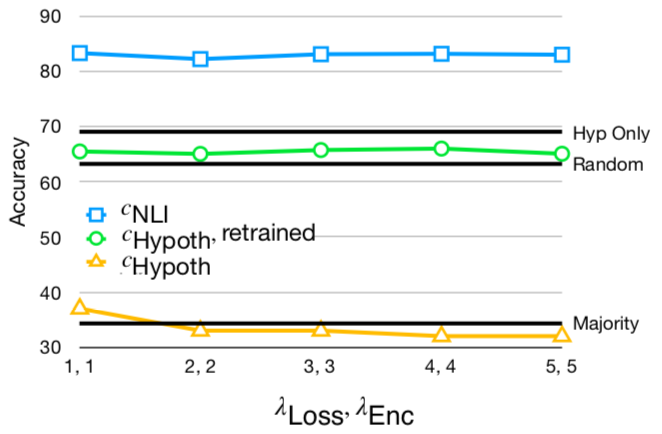}
\caption{Hidden biases remaining from \HOC}
\label{fig:frozen-double}
\vspace{10pt}
\end{subfigure} \\ 
\begin{subfigure}[t]{\linewidth}
\centering
\includegraphics[width=\linewidth,trim={0 0 0 1.3cm},clip]{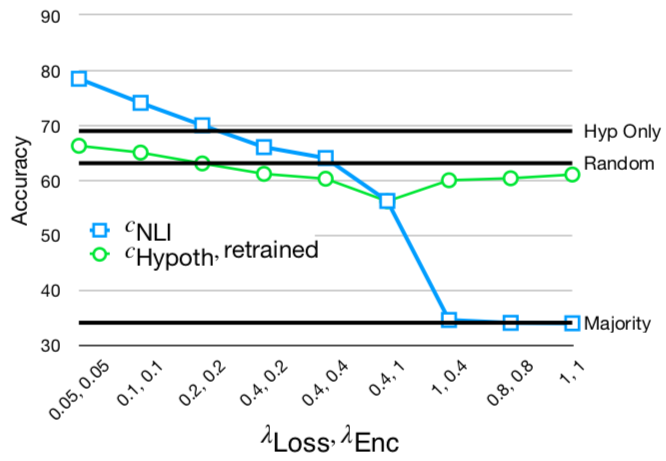} 
\caption{Hidden biases remaining from \NS}
\label{fig:frozen-single}
\vspace{5pt}
\end{subfigure}
\caption{Validation  
results when
retraining a 
classifier on a frozen hypothesis encoder ($\classifier_\text{Hypoth}$, retrained) compared to our methods ($\classifier_\text{NLI}$), the 
adversarial hypothesis-only classifier ($\classifier_\text{Hypoth}$, in \HOC),
majority baseline, a random frozen encoder, and a hypothesis-only
model.
} 
\label{fig:frozen}
\vspace{-5pt}
\end{figure}

\subsection{Hidden Biases}

\begin{figure*}[t]
\centering
\begin{subfigure}[t]{.48\linewidth}
\centering
\includegraphics[width=\linewidth,trim={0 0 0 .3cm},clip]{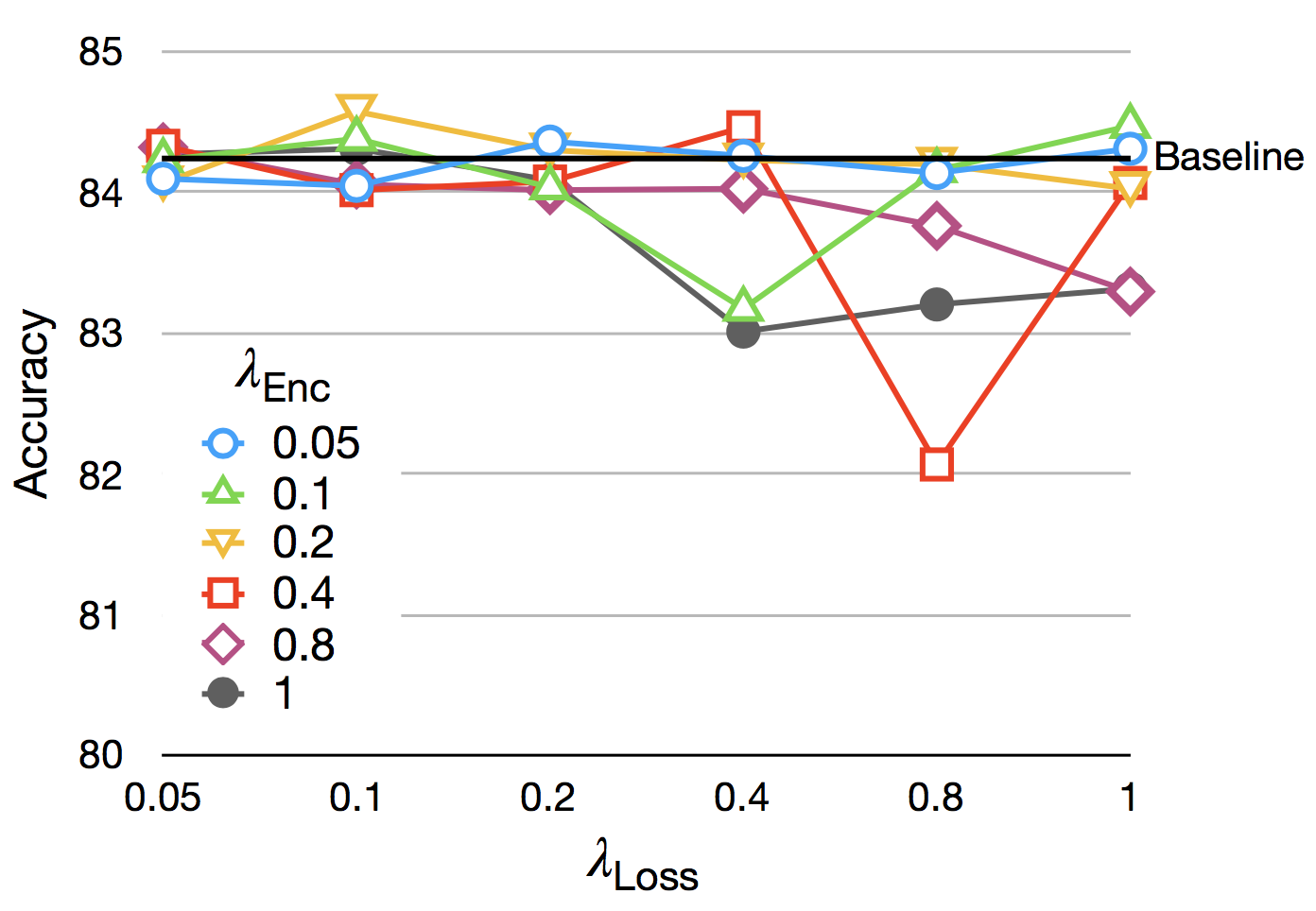}
\caption{\HOC}
\label{fig:hparams-double}
\end{subfigure} \hfill
\begin{subfigure}[t]{.48\linewidth}
\centering
\includegraphics[width=\linewidth,trim={0 0 0 .3cm},clip]{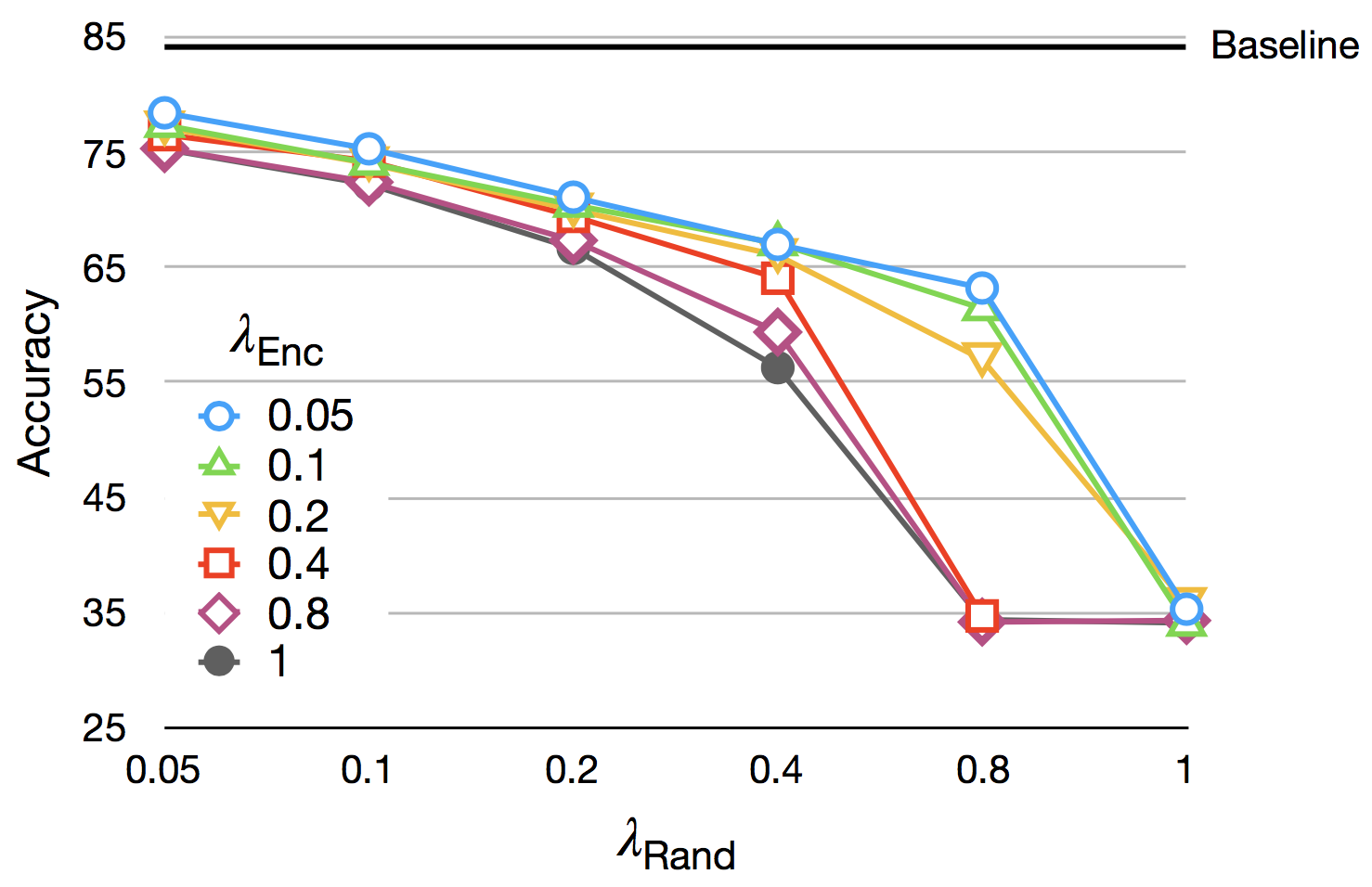} 
\caption{ \NS}
\label{fig:hparams-single}
\end{subfigure}
\caption{Results on the validation set with different configurations of the adversarial methods.}
\label{fig:hparams}
\vspace{-5pt}
\end{figure*}

Do the learned 
sentence representations eliminate hypothesis-only biases after adversarial training?
We freeze sentence encoders trained with the studied methods,
and retrain a new classifier that only
accesses representations
from the frozen hypothesis encoder.
This helps us determine whether the (frozen) 
representations have hidden biases.

A few trends can be noticed. First,
we confirm that with \HOC 
(\newfigref{fig:frozen-double}), the hypothesis-only classifier ($\classifier_\text{hypoth}$) is indeed trained to perform poorly on the task, 
while the normal NLI classifier ($\classifier_\text{NLI}$)
performs much better. 
However, retraining a classifier  on frozen hypothesis representations ($\classifier_\text{Hypoth}$, retrained)
boosts 
performance. In fact, the retrained classifier performs close to the fully trained hypothesis-only baseline, 
indicating 
the hypothesis representations still contain
biases. 
Consistent with this finding,
~\citet{elazar2018adversarial} found that adversarially-trained text 
classifiers preserve 
demographic attributes in 
hidden representations despite efforts to remove them.

\begin{table*}[t]
\centering
\begin{tabular}{l r  rrrrr}
\toprule
& & \multicolumn{2}{c}{Score} & \multicolumn{3}{c}{Percentage decrease 
from baseline
} \\
\cmidrule(lr){3-4} \cmidrule(lr){5-7}
Word & Count & 
$\hat{p}(l|w)$& Baseline
& \HOC (1,1) & \NS (0.4,1) & \NS (1,1) \\
\midrule
sleeping & 108 & 0.88 & 0.24 & 15.63 & 53.13 & -81.25 \\
driving & 53 & 0.81 & 0.32 & -8.33 & 50 & -66.67 \\
Nobody & 52 & 1 & 0.42 & 14.29 & 42.86 & 14.29 \\
alone & 50 & 0.9 & 0.32 & 0 & 83.33 & 0 \\
cat & 49 & 0.84 & 0.31 & 7.14 & 57.14 & -85.71 \\
asleep & 43 & 0.91 & 0.39 & -18.75 & 50 & 12.5 \\
no & 31 & 0.84 & 0.36 & 0 & 52.94 & -52.94 \\
empty & 28 & 0.93 & 0.3 & -16.67 & 83.33 & -16.67 \\
eats & 24 & 0.83 & 0.3 & 37.5 & 87.5 & -25 \\
naked & 20 & 0.95 & 0.46 & 0 & 83.33 & -33.33 \\
\bottomrule
\end{tabular}
\caption{Indicator words and how correlated they are with  \textsc{CONTRADICTION} predictions. The parentheses indicate hyper-parameter values: $(\lambda_\text{Loss},\lambda_\text{Enc})$ for \HOC and $(\lambda_\text{Rand},\lambda_\text{Enc})$ for \NS.
Baseline refers to the unmodified 
InferSent.
}
\label{tab:indicators}
\end{table*}

Interestingly, we found that even a frozen random encoder captures biases in the hypothesis, as a classifier trained on it performs fairly well (63.26\%), and far above the majority class baseline (34.28\%). 
One reason might be that the word embeddings (which are pre-trained) alone contain significant information that propagates even through a random encoder. Others have also found that random encodings contain non-trivial information~\citep{conneau2018you,zhang2018language}.
The fact that the word embeddings were not updated during (adversarial) training could account for the ability to recover performance at the level of the classifier trained on a random encoder.
This may indicate that future adversarial efforts should be applied
to the word embeddings as well.

Turning to 
\NS, (\newfigref{fig:frozen-single}),
as the hyper-parameters increase,
the models exhibit fewer biases. Performance even drops below the random encoder results, indicating 
it
may be better at
ignoring
biases in the hypothesis. However, this comes at the cost of reduced
NLI performance.

\subsection{Adversarial Strength} \label{sec:strength}
Is there a correlation between adversarial strength and drops in SNLI performance? 
Does increasing adversarial hyper-parameters affect the decrease in results on SNLI?

\newfigref{fig:hparams} shows the validation results with various configurations of adversarial hyper-parameters. The \HOC method is fairly stable across configurations, although combinations of large $\lambda_\text{Loss}$ and $\lambda_\text{Enc}$ hurt the performance
on SNLI a bit more (\newfigref{fig:hparams-double}). Nevertheless, all the drops are moderate. Increasing the hyper-parameters further (
up to values of $5$), did not lead to substantial drops, although the results are slightly less stable across configurations (\appref{app:hparams-stronger}).
On the other hand, the \NS method is very sensitive to large hyper-parameters (\newfigref{fig:hparams-single}). For every value of $\lambda_\text{Enc}$, increasing $\lambda_\text{Rand}$ leads to significant performance drops. These drops happen sooner for larger 
$\lambda_\text{Enc}$ values.
Therefore,
the effect of
stronger hyper-parameters on 
SNLI performance  
seems to be specific to each adversarial method.

\subsection{Indicator Words}
\label{sec:indicators}

Certain words in SNLI are more correlated with specific entailment labels than others,
e.g., negation words (``not'', ``nobody'', ``no'') correlated with \textsc{contradiction}~\citep{gururangan-EtAl:2018:N18-2,poliak-EtAl:2018:S18-2}.
These words have been referred to as ``give-away''
words~\cite{poliak-EtAl:2018:S18-2}.
Do the adversarial
methods encourage models to make predictions that are less affected by these biased indicator words? 

For each of the most biased words in SNLI associated with the \textsc{contradiction} label, 
we computed the probability that a  model 
predicts an example as a contradiction, given that the hypothesis contains the word.
\tabref{tab:indicators} shows the top 10 
examples in the training set. 
For each word $w$, we give its frequency in SNLI, its empirical correlation 
with the label and
with 
InferSent's prediction, and the percentage decrease 
in correlations with \textsc{contradiction} predictions by three configurations of our methods.
Generally,
the baseline correlations are more uniform than the empirical ones (\^{p}$(l|w)$),
suggesting that indicator words in SNLI might not
greatly affect a
NLI model, 
a possibility that both \citet{poliak-EtAl:2018:S18-2} and \citet{gururangan-EtAl:2018:N18-2} do
concede.
For example, \citet{gururangan-EtAl:2018:N18-2} explicitly mention that ``it is important to note
that even the most discriminative words are not
very frequent.''

However, we 
still observed small 
skews towards \textsc{contradiction}.
Thus, we investigate whether our methods reduce the probability of predicting \textsc{contradiction} 
when a hypothesis contains an indicator word. The model trained with \NS
(where $\lambda_\text{Rand} = 0.4$, $\lambda_\text{Enc} = 1$) 
predicts contradiction much less frequently than
InferSent
on examples with these words. This configuration was the strongest 
\NS 
model that still performed reasonably well on SNLI (\newfigref{fig:hparams-single}). 
Here, 
\NS 
appears to 
remove some of the biases learned by the baseline, unmodified InferSent. 
We also provide two other 
configurations that  do not show such an effect, 
illustrating that this behavior highly depends on the 
hyper-parameters.

\section{Conclusion}

We employed two adversarial learning techniques to a general NLI model by adding an external adversarial hypothesis-only classifier and perturbing training examples.
Our experiments and analyses suggest that these techniques may help
models exhibit fewer
hypothesis-only biases.
We hope 
this work will encourage the development and analysis of 
models that include components that ignore hypothesis-only biases, as well as 
similar biases discovered in other natural language understanding 
tasks~\cite{schwartz2017effect},
including visual question answering, where recent work has considered similar adversarial techniques for removing language biases~\cite{ramakrishnan2018overcoming,grand:2019:SIVL}.

\section{Acknowledgements}
This work was supported by JHU-HLTCOE,
DARPA LORELEI, the Harvard Mind, Brain, and Behavior Initiative, and NSF 1845664.
We thank the anonymous reviewers for their comments.
Views and conclusions contained in this publication are those of the authors and should not be interpreted as representing official policies or endorsements of DARPA or the U.S.\ Government.

\bibliography{references}
\bibliographystyle{iclr2019_conference}

\appendix

\newpage
\section{Stronger hyper-parameters for \HOC }  \label{app:hparams-stronger}
\begin{figure}[h]
\centering
\includegraphics[width=\linewidth,trim={0 0 0 1.5cm},clip]{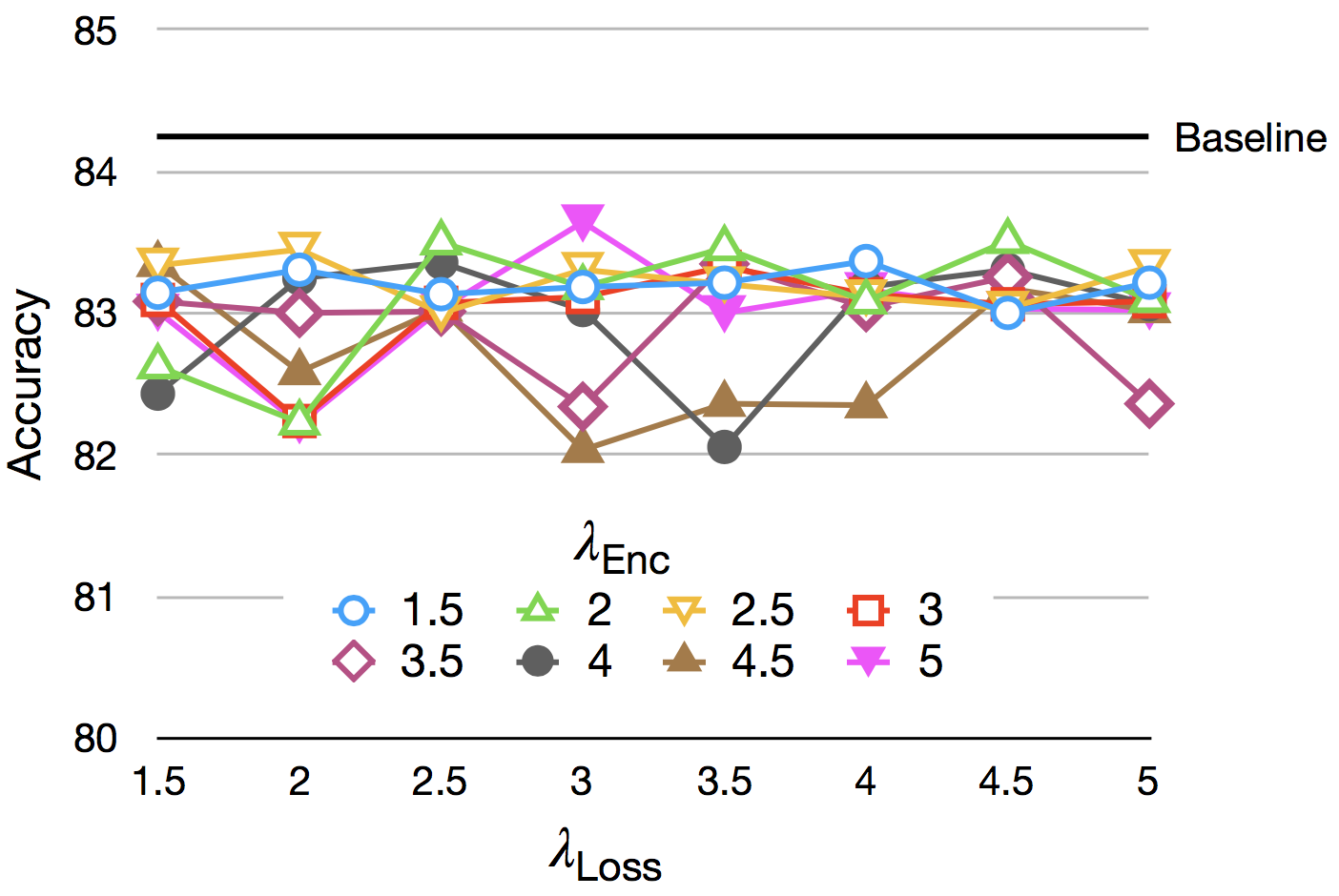}
\caption{Validation results using \HOC with stronger hyper-parameters.}
\label{fig:hparams-double-stronger}
\end{figure}

\newfigref{fig:hparams-double-stronger} provides validation results using \HOC with stronger hyper-parameters to complement the discussion in \newsecref{sec:strength}. While it is difficult to notice trends, all configurations perform similarly and slightly below the baseline. These models seem to be less stable compared to using smaller hyper-parameters, as discussed in \newsecref{sec:strength}.

\end{document}

%% file: math_commands.tex
\usepackage{amsmath,amsfonts,bm}

\def\eqref#1{equation~\ref{#1}}

\def\1{\bm{1}}

\def\vh{{\bm{h}}}

\def\vp{{\bm{p}}}

\def\vv{{\bm{v}}}

\DeclareMathAlphabet{\mathsfit}{\encodingdefault}{\sfdefault}{m}{sl}
\SetMathAlphabet{\mathsfit}{bold}{\encodingdefault}{\sfdefault}{bx}{n}

